\documentclass[
]{ceurart}

\usepackage{listings}

\usepackage{graphicx}
\usepackage{tcolorbox}
\tcbuselibrary{listingsutf8}
\usepackage{changepage}
\usepackage{alltt}
\usepackage{verbatim}
\usepackage{tabularx}
\usepackage{booktabs}
\usepackage{float}

\usepackage{subcaption}
\begin{document}

\copyrightyear{2025}
\copyrightclause{Copyright for this paper by its authors.
  Use permitted under Creative Commons License Attribution 4.0
  International (CC BY 4.0).}

\conference{IberLEF 2025, September 2025, Zaragoza, Spain}

%%\title{GRESEL Team at PastReader 2025: Experiments with Transkribus, Tesseract and Granite}

\title{Transcribing Spanish Texts from the Past: Experiments with Transkribus, Tesseract and Granite}
\author[1]{Yanco Amor Torterolo-Orta}[%
orcid=0000-0002-3688-3293,
email=ytorterolo@lsi.uned.es,
%url=https://yamadharma.github.io/,
]
\cormark[1]
\fnmark[1]

\author[1]{Jaione Macicior-Mitxelena}[%
orcid=0009-0001-7392-4226,
email=jmacicior2@alumno.uned.es,
%url=https://kmitd.github.io/ilaria/,
]
\fnmark[1]

\author[1]{Marina Miguez-Lamanuzzi}[%
orcid=0000-0002-1941-8031,
email=marimigu@ucm.es,
%url=http://conceptbase.sourceforge.net/mjf/,
]
\fnmark[1]

\author[1]{Ana García-Serrano}[%
orcid=0000-0003-0975-7205,
email=agarcia@lsi.uned.es,
%url=http://conceptbase.sourceforge.net/mjf/,
]
\fnmark[1]

\address[1]{Universidad Nacional de Educación a Distancia (UNED),
  Madrid, Spain}

\cortext[1]{Corresponding author.}
\fntext[1]{These authors contributed equally.}

\begin{abstract}
This article presents the experiments and results obtained by the GRESEL team in the IberLEF 2025 shared task \textit{PastReader: Transcribing Texts from the Past}. Three types of experiments were conducted with the dual aim of participating in the task and enabling comparisons across different approaches. These included the use of a web-based OCR service, a traditional OCR engine, and a compact multimodal model. 
All experiments were run on consumer-grade hardware, which, despite lacking high-performance computing capacity, provided sufficient storage and stability.
The results, while satisfactory, leave room for further improvement. Future work will focus on exploring new techniques and ideas using the Spanish-language dataset provided by the shared task, in collaboration with Biblioteca Nacional de España (BNE).

\end{abstract}


\maketitle

\section{Introduction}

Documents from the past have much to say in the present, just as history teaches people so that it does not repeat itself. However, old documents with significant historical and cultural value can be challenging to use computationally, hindering efforts to extract information from them. This is why the shared task \textit{PastReader: Transcribing Texts from the Past} \cite{MontejoRaez2025} is so relevant. This shared task was organized as part of the 2025 edition of the Iberian Languages Evaluation Forum (IberLEF 2025) \cite{iberlef2025overview}. This paper describes the joint participation of  two teams, GRESEL1 \cite{gresel1} and GRESEL2 \cite{gresel2}. GRESEL (AI generation results enriched with simplified explanations based on linguistic features) is a Spanish National Project (PID2023-151280OB) devoted to the development of new domain-specific resources and adapting existing linguistic tools and methodologies to extract relevant information from the corpora provided by project participants. This includes training language models (LMs) tailored to the specific needs of the project, identifying and extracting meaningful words (e.g., financial terms, place names, informative words in a novel) and categorising them according to different domains (musical instruments, accounting concepts, suffragism and the role of women, colonial interactions, etc.) to cater to the diverse interests of potential users. 

The GRESEL team at PastReader shared task is highly multidisciplinary, with philologists, computational linguists, and computer scientists. Previous work of the team includes approaches for the reusability of existing models through fine-tuning and transfer learning techniques, also when the available corpus is written in a language or domain with scarce resources such as the digitised collection of newspapers called \textit{Diario de Madrid} (DM) from the Spanish press between 18th and 19th centuries, which is freely available at the Spanish National Library (BNE) \cite{sanchez2023seeking}; open benchmarks for new technologies \cite{LASTRA-2021} or the work in Image Captioning from related texts \cite{SEPLN2025}. Also, some members of the group worked in how to leverage multimodal approaches for different tasks in multimedia corpus \cite{clef-2008, clef-2005}. Some other related contributions include the organisation of \textit{The Financial Document Causality Detection Shared Task (FinCausal 2025)} \cite{fincausal}, and the best-scoring participation at the shared task \textit{FinancES 2023: Financial Targeted Sentiment Analysis in Spanish} \cite{finances}. Additionally, RAG experiments on literary works are being conducted \cite{sepln-Yanco}.

Given the similarities observed in both PastReader tasks and the provided corpus (Train, Development, and Test) with the Gresel team's previous experience, as well as the nature of the shared task, there is a strong alignment in goals. Therefore, we decided to participate with the aim of comparing previously used systems with new approaches. Achieving the highest scoring system was not a goal, but a way to share with the community our findings.

We designed three different approaches. The first one is based on the use of Transkribus\footnote{\url{https://www.transkribus.org/}}, an AI-based platform offering various functionalities accessible through its web application. It is an excellent choice for non-expert users in technology who wish to annotate documents for fine-tuning Optical Character Recognition (OCR) models. The second and third approaches are based on currently available open models, such as Tesseract, supported by Google\footnote{\url{https://sourceforge.net/projects/tesseract-ocr.mirror/}}; and Granite, created by IBM\footnote{\url{https://www.ibm.com/es-es/granite}}.

Historically, OCR has always been one of the main topics of research and development given its real-world applications. In fact, it is increasingly drawing attention thanks to recent advancements in multimodal models, both open-source and proprietary. \textit{Llava} \cite{llava} is a well-known multimodal model, but some open-source models that provide a multimodal version include Meta's \textit{Llama3.2-vision} \cite{llama} or Google's \textit{Gemma3-vision} \cite{gemma}. On the proprietary side, OpenAI offers several multimodal models like \textit{GPT-4-turbo}, \textit{GPT-o4-mini} or \textit{GPT-o3}\footnote{\url{https://platform.openai.com/docs/models/compare}}. Currently, there is a vast number of research leveraging multimodal models for OCR. One example is \cite{olmocr}, with an open-source toolkit available for people to fine-tune their models on English OCR tasks. They provide both the dataset used, \textit{olmOCR-mix-0225}, and their fine-tuned model, \textit{olmOCR-7B-0225-preview}, which is based on \textit{Qwen2-VL-7B-Instruct} \cite{qwen}.

It is worth noting that the emergence of the Digital Humanities and the growing awareness of the importance of preserving and digitising analogue data for future use make this task particularly relevant.

The rest of this work is structured as follows: Section 2 contextualises the shared task, introduces our main approaches, and analyses the dataset, highlighting certain aspects that hinder model performance. Section 3 describes our three main approaches in detail, and Section 4 provides an in-depth analysis of the results. Some conclusions are drawn at the end in Section 5, offering insights into the task and directions for future work.

\section{Task description and approaches}

PastReader 2025 consists of applying OCR on PDF files with limited quality. The dataset is described in detail below. Two different, yet related, tasks were proposed. On the one hand, OCR outputs from said PDF files are provided. These outputs are in plain text and purposely exhibit a suboptimal quality. Hence, the task 1 is to provide clean versions of these texts. The clean texts are meant to align with the ground truth. On the second hand, the task 2 requires developing an end-to-end OCR system. PDF files are the intended input, ignoring the proposed OCR text outputs from task 1. From the PDF files, new text outputs are expected to be provided directly. The main challenge lies in the dataset quality and the variety of the sources.

As Transkribus application was explored in some of our previous work, and given that time constrains were a factor to be considered for the three approaches, we decided to participate only in task 2. As anticipated, our participation consists of three different approaches, the remaining ones being Tesseract and \textit{Granite3.2-vision:2b} \cite{granite}. These three approaches are quite different and represent different OCR paradigms.

\begin{itemize}
    \item \textbf{Transkribus} is a web application that uses AI models for text recognition and for the training of AI models aimed at the transcription and recognition of texts. This application has useful features for text processing, such as the ability to annotate the different parts of a document (header, body text, footer notes, etc.) and to label the different parts of the documents for the AI models' training. It is an application specifically designed for processing historical texts, as is the case with our dataset. It is a platform that also encourages teamwork, since it is possible to create projects with the collaboration of different users and jointly edit the collections of documents. We used Transkribus in the task to test how the \textit{Coloso Español}\footnote{\url{https://readcoop.eu/model/coloso-espanol}} model transcribes the dataset documents. \textit{Coloso Español} is a model specifically trained to recognise historical Spanish texts. 
    
    \item \textbf{Tesseract} has also been applied to process historical documents. Its open-source nature and adaptability make it a practical choice for initial OCR tasks on printed texts including historical documents. Tesseract was developed by Hewlett-Packard and later maintained by Google. It supports over 100 languages and includes features such as layout analysis and recognition of various image formats. 

    \hspace*{1em}However, Tesseract's performance can be challenged when dealing with historical documents that often contain noise, complex layouts, or uncommon fonts. A benchmarking study comparing Tesseract with Amazon Textract and Google Document AI found that server-based processors outperformed Tesseract, especially on noisy documents \cite{hegghammer2022ocr}. Despite these limitations, Tesseract remains a valuable tool for establishing baseline OCR performance and serves as an useful point of comparison when evaluating more advanced or specialised OCR systems.

    \hspace*{1em}Recent research has explored enhancing Tesseract's capabilities for historical documents. For instance, a study focused on 19th-century printed serials employed a combined machine learning approach, fine-tuning Tesseract with synthesised and manually annotated data to improve OCR quality \cite{fleischhacker2024improving}. This approach demonstrated that, with appropriate training, Tesseract's accuracy on historical texts could be significantly enhanced. 

    \item \textbf{Granite3.2-vision:2b} was selected over other open-source multimodal models for two main reasons. The first one is that we value the democratisation of LLMs and being able to fine-tune small models on consumer-grade hardware. This is a small model, with only 2 billion parameters, which raised the chances of it fitting in the 16gb of VRAM of the Nvidia RTX 5080 employed for this matter. The second reason is the high performance this model offers despite its size.

    \hspace*{1em}According to IBM\footnote{\url{https://www.ibm.com/new/announcements/ibm-granite-3-2-open-source-reasoning-and-vision}}, this model focuses on being especially performant for enterprises. It has been fine-tuned on IBM's DocFM dataset. It is a ``large instruction tuning dataset'' consisting of high-quality enterprise data. Basically, it prioritises visual document understanding with both image and text, which encompasses document characteristics such as layouts, fonts, charts, etc. Conversely, other models are mainly trained with natural images, allegedly yielding worse results than IBM's model. In fact, based on IBM's claims, this model rivals larger models in benchmarks such as DocVQA and ChartQA. Granite stands out in document understanding and multimodal retrieval-augmented generation (RAG). The remarkable ability of multimodal models to ``understand'' images and answer questions about said images makes them a perfect choice for OCR. Therefore, it was considered a suitable choice for this shared task.

\end{itemize}

\subsection{Datasets analysis}

The dataset \cite{MontejoRaez2025dataset} provided for this shared task consisted of historical documents sourced from various archives and repositories. As a result, it was highly heterogeneous in nature, including variations in font styles, print quality, background colouration, presence of handwritten marginalia, and embedded visual elements such as stamps or illustrations. This diversity closely reflects the real-world conditions under which digitisation efforts must operate, but it also presents a significant challenge for OCR systems that are typically trained on more standardised inputs.

These inconsistencies in the visual presentation of the documents have a direct impact on OCR performance. Models like Tesseract, which rely heavily on traditional pattern recognition and rule-based layout analysis, often struggle with non-uniform inputs. Differences in font type and noise introduced by paper degradation or scanning artefacts can significantly increase the character error rate (CER) and word error rate (WER). For example, documents with faded ink or coloured backgrounds may result in missed characters or misclassification, particularly when the contrast between text and background is insufficient. 

Transkribus, which uses machine learning and can be trained on specific document types, tends to handle variability more effectively. Its performance benefits from custom model training and a more flexible recognition pipeline. However, the generalisation ability of a model trained on one subset of documents may still decline when faced with very different styles from another subset. This is especially problematic in historical corpora, where typographic standards and orthographic conventions were far from unified.

Granite, being a multimodal model, demonstrates a different set of strengths and weaknesses. Its broader contextual understanding and vision-language architecture offer promise for interpreting complex layouts and mixed media. However, like any foundation model, its effectiveness is highly dependent on how similar the dataset is to its pretraining data. The wide variation in the dataset can either help by diversifying test conditions or hinder performance if the model has not seen enough similar samples during training. These results underline the importance of dataset consistency or adaptive fine-tuning when deploying OCR systems in heterogeneous historical collections.

\subsection{Some interesting aspects of the Dataset}
\label{subsec:complexities}

The provided dataset is composed by isolated pages of 8 periodical publications from the 19th and 20th centuries in Spanish language. The topics they cover are very varied, assorted from cultural and general news magazines (\textit{Juventud}, \textit{El Español}), to satirical and humorous magazines (\textit{El Duende Satírico del día}), chronicles of spiritualism (\textit{La luz del porvenir}), serial publications of narrative fiction and poetry (\textit{Revista nueva}, \textit{La patria de Cervantes}) or even newspapers about fashion and feminine customs (\textit{Periódico de las damas}), as well as scientific publications on medical issues (\textit{Revista frenopática española}). As observed, these press pages differ significantly in their goals, themes, and formats.

The following are examples of pages to highlight several complexities that models should be designed to handle. A dedicated model should be trained with these and other common challenges found in historical Spanish texts in mind, as contemporary texts are not a suitable substitute. However, due to time constraints, the GRESEL team has not yet developed such a model using any of the three explored approaches, as creating a dataset with a sufficient number of representative examples remains pending.

\begin{figure}[ht]
\centering
\begin{tabular}{>{\centering\arraybackslash}m{0.32\textwidth}
                >{\centering\arraybackslash}m{0.32\textwidth}
                >{\centering\arraybackslash}m{0.32\textwidth}}
    \begin{minipage}[t]{\linewidth}
        \centering
        \includegraphics[width=0.95\linewidth]{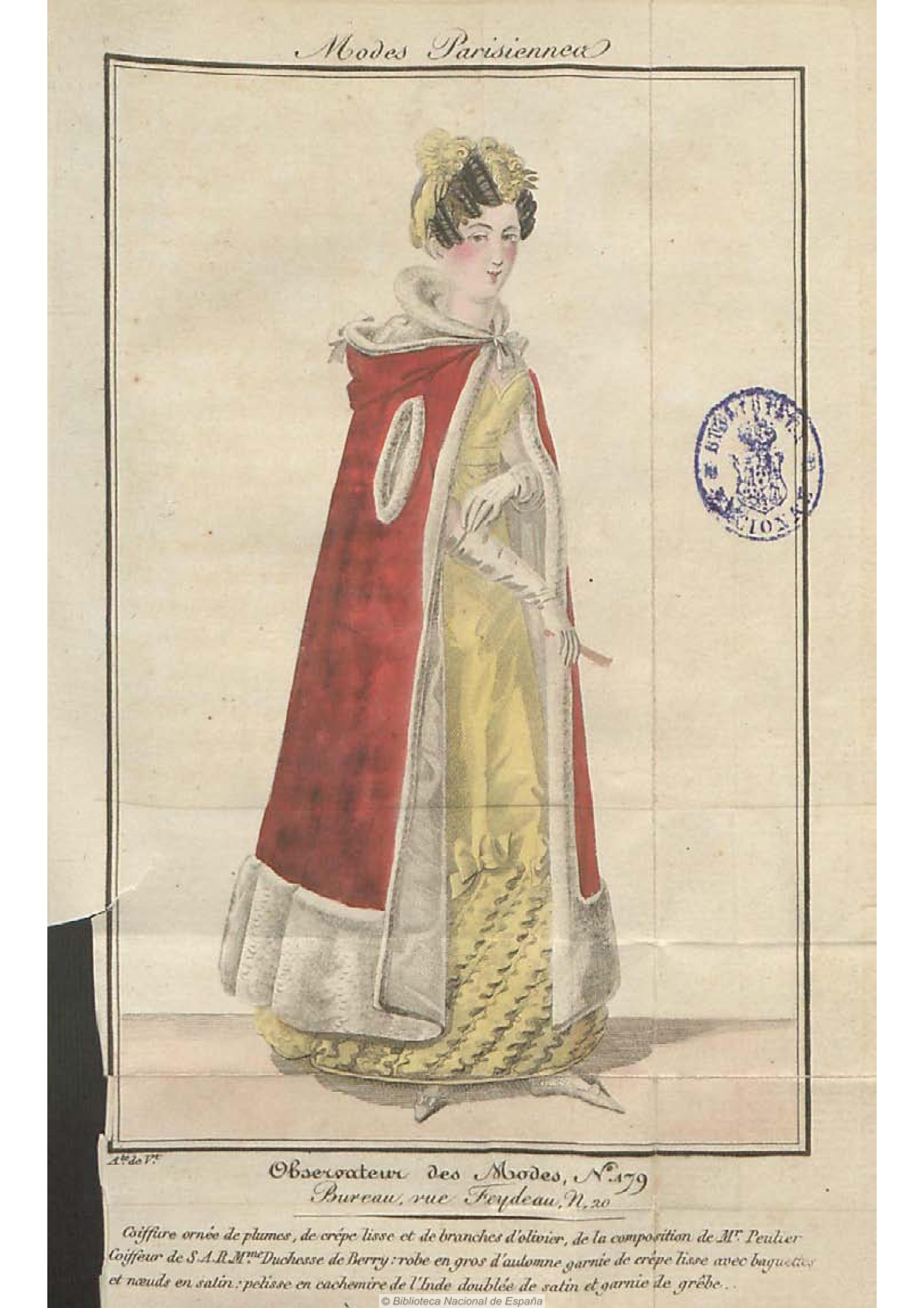}
        \par Page 9062
        \label{fig:9062}
    \end{minipage} &
    \begin{minipage}[t]{\linewidth}
        \centering
        \includegraphics[width=0.95\linewidth]{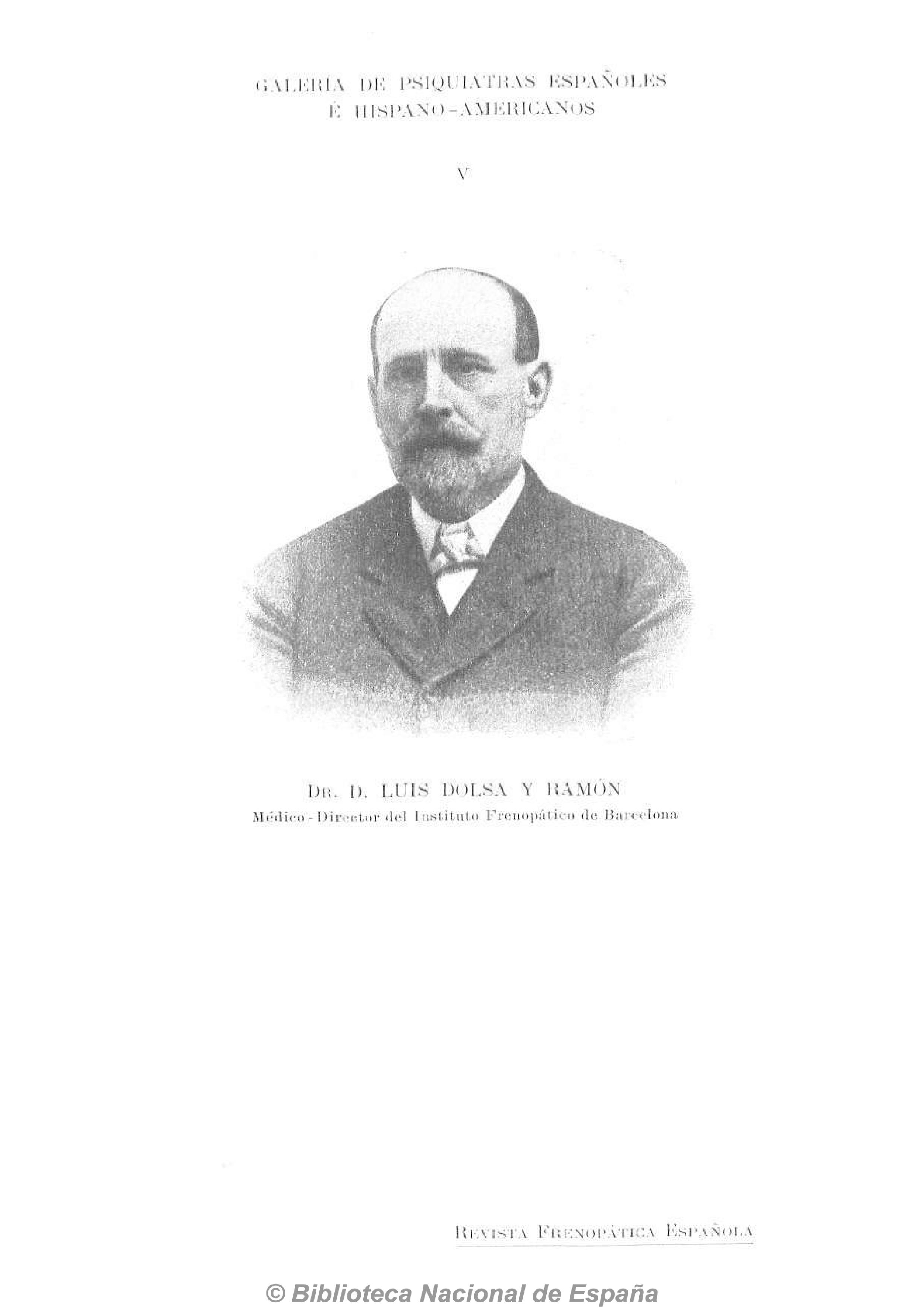}
        \par Page 9171
        \label{fig:9171}
    \end{minipage} &
    \begin{minipage}[t]{\linewidth}
        \centering
        \includegraphics[width=0.95\linewidth]{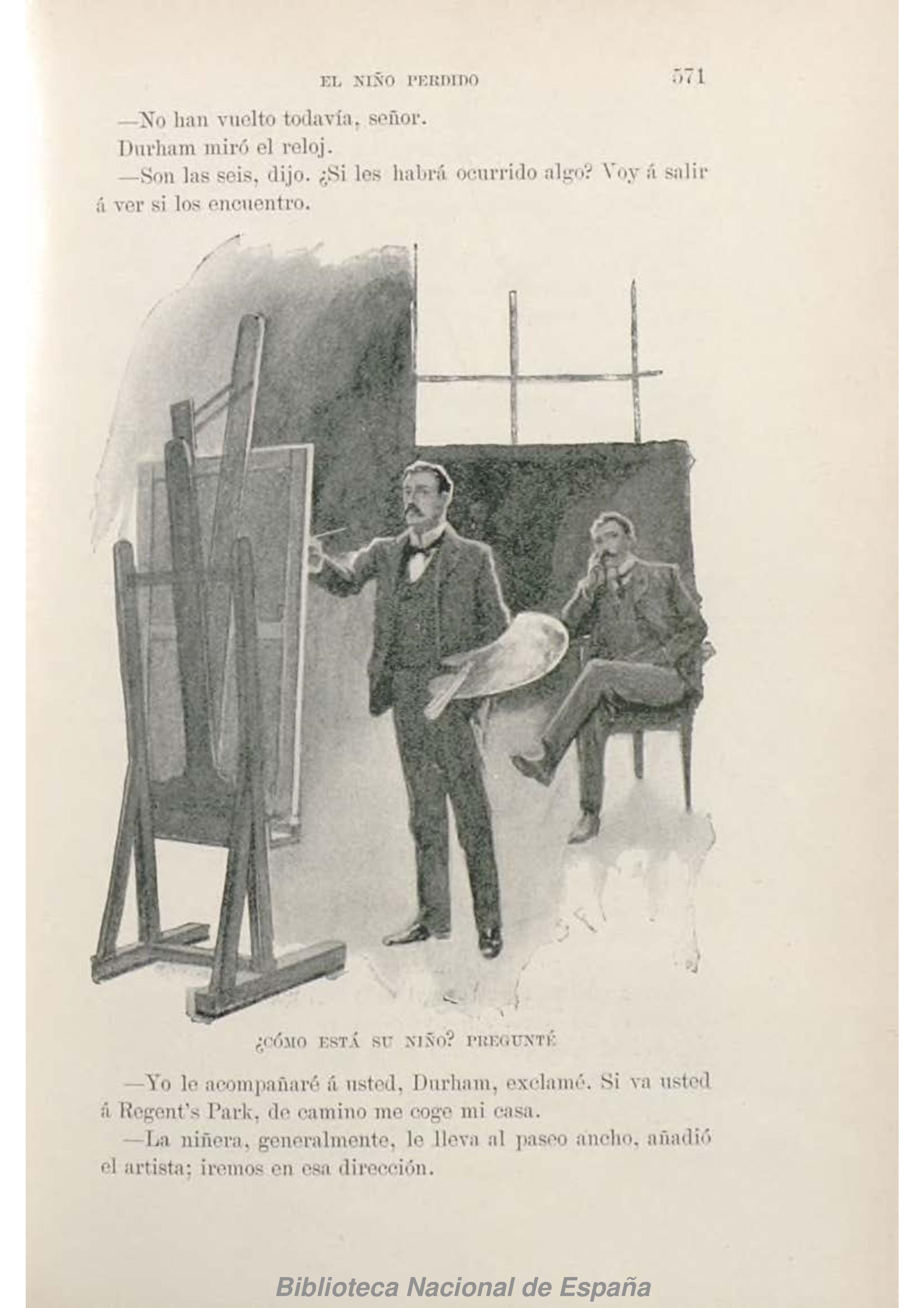}
        \par Page 8963
        \label{fig:8963}
    \end{minipage} \\
\end{tabular}
\caption{Examples of pages with poor performance.}
\label{fig:worst_pages}
\end{figure}

First of all we can see that many of the examined documents have in common that they combine images and text on the page, for the most part narrative (novels, short stories), where the image is simply a complement to the narration, to provide it with greater expressive vividness. However, in other cases, there are documents where the image is the main element, and the text is simply a complement to it. For example, this is the case of a page from a fashion magazine (9062, Figure \ref{fig:worst_pages}), where the text associated with the image is the description of the outfit; in other cases, it is the portrait of a phrenologist doctor (9171, Figure \ref{fig:worst_pages}), where the associated text is just the doctor's name and their speciality. Thus, it is necessary to take into account that the hierarchy of information is reversed in these two cases, with respect to the informative importance that the images have in relation to the text.

Also, in many of the examined pages, we find that the image, and a brief text associated with it in small capitals, interrupts the transcription of the sentences of the narrative texts (8963, Figure \ref{fig:worst_pages}). In fact, paratextual elements such as image captions are expected to disrupt the typical linear reading order followed by OCR systems—from top to bottom, left to right. Another aspect that can hinder the OCR is the presence of typographical ornamentation, like small and decorated capitals. This phenomenon can be observed in the examples above. This can potentially lead the models to misinterpret some letters.

%Another issue is that, in many of the examined pages, we find that the image, and a brief text associated with it in small capitals, interrupts the transcription of the sentences of the narrative texts (8963, Figure \ref{fig:8963}). In fact, the \textit{Coloso Español} model transcribes the small capital text associated with the images when they encounter it during their ``reading.'' This causes a lot of confusion in the transcriptions, which are interrupted by paratextual elements such as image comments.

%Another important issue is that the model does not detect indentation, which is what indicates a paragraph change in the text. Instead, it separates the text into lines exactly as it finds them on the page, which does not respect textual cohesion and coherence.

%Finally, the Transkribus \textit{Coloso Español} model (without checking the other two approaches) does not correctly interpret typographical issues, such as small and decorated capitals. Furthermore, it confuses some characters, such as the letters ‘o’ and ‘c,’ with the letter ‘e,’ or interprets what are actually part of illustrations or printing marks as characters. 

In order to obtain a good training of AI models and reliable results, we consider that it is essential that the dataset is clean and well organised, as well as the application of systematic transcription rules based on coherent choices from a philological and palaeographic point of view. For this reason, we are preparing annotation and transcription guidelines based on the issues detected in the documents of this training dataset for future work.

\section{Experiments}

Our three main experiments are described in the following lines. The first one, Transkribus, was used for inference using an existing model for Spanish due to a lack of time for training a model. The other two experiments include fine-tuning on the dataset provided by the competition.

\subsection{Transkribus}
In this approach, all documents in the test corpus were processed using one of the models available in the Spanish application, specifically, as mentioned earlier, \textit{Coloso Español}. It was trained in collaboration with various researchers using Spanish models for historical texts from different periods, from medieval manuscripts to 20th-century documents.
It was necessary to use Transkribus paid credits and export the results in two formats, with and without line breaks (\textbf{GRESEL1\_run1} and \textbf{GRESEL2\_run2}, respectively), to finally configure the TXT files in the format required by PastReader shared tasks.
As already mentioned, it would be worth training a Transkribus model from a higher quality corpus (approximately 60 pages) that included semi-manual transcriptions without errors coping the complexities identified in Section \ref{subsec:complexities}, but it goes beyond the scope of this paper.

\subsection{Tesseract}

Tesseract, as mentioned earlier, is an open-source OCR engine\footnote{\url{https://github.com/tesseract-ocr/tesseract}}. The installation was performed on a Windows-based system, but it can be done in other operating systems easily. Fine-tuning was implemented using the tesstrain training toolkit, which is compatible with the LSTM-based recognition architecture introduced in Tesseract 4.0 and maintained in later versions.

Two experimental runs were conducted to assess Tesseract’s performance under different configurations. In the first run \textbf{GRESEL1\_run2}, the default pre-trained Tesseract model was applied directly to the test set without any additional training or adaptation. This approach simulates a zero-shot or ``off-the-shelf'' application scenario, which is useful for establishing a baseline. The model was executed with standard parameters, and no preprocessing was applied beyond resizing to ensure compatibility with Tesseract’s input expectations. In the second run \textbf{GRESEL1\_run3}, the baseline was fine-tuned on the provided training dataset to tailor it more closely to the characteristics of the historical documents. The purpose of this fine-tuning was to determine whether training on similar materials could improve recognition accuracy—especially on degraded, low-contrast, or typographically irregular documents that typically reduce Tesseract's performance.

This two-step process was not only useful for benchmarking the Tesseract model but also served as a practical diagnostic tool for assessing the quality of the dataset itself. By observing the kinds of errors the model made—both before and after fine-tuning—, it was possible to identify problematic samples, inconsistencies in labelling, and areas where document quality or transcription standards might affect downstream OCR performance. These insights were valuable for both improving model training and better understanding the dataset's challenges and limitations. 

%    Training process ( GRESEL1\_run3) 

The fine-tuning was conducted using version 5.5.0 of Tesseract, installed with training support enabled (by selecting the ``Install training tools'' option during setup). The official Windows installer with training tools was downloaded from the Tesseract GitHub releases page. All scripts and auxiliary tools were developed and executed in a Windows environment. The tesstrain toolkit was cloned from the official GitHub repository\footnote{\url{https://github.com/tesseract-ocr/tesstrain.git)}} to provide the core infrastructure for training. A base Spanish model (spa.traineddata) was manually placed in a custom tessdata/ directory, and used as the initialisation point for transfer learning. To streamline the process, a Python script (Finetune\_tesseract.py) was developed to automate the full fine-tuning pipeline. The structure of the working directory before fine-tuning was organised as described in Figure \ref{fig:working-dir-combined} to the left. 

\begin{figure}[htbp]
\centering
\begin{tcolorbox}[
    colback=white,
    colframe=black,
    boxrule=0.5pt,
    width=\textwidth,
    title=WORKING DIRECTORY STRUCTURE
]
\begin{tabular}{p{0.48\textwidth} p{0.48\textwidth}}
\textbf{Before Fine-Tuning} & \textbf{After Fine-Tuning} \\
\begin{adjustwidth}{0em}{0pt}
\begin{alltt}
\normalfont
PastReader/ 
|-- train/ 
|   |-- pdf/     \textit{<- Original PDF documents ->}                
|   |__ ocr/    \textit{<- Transcripts (.txt) ->}
|
|
|-- finetuning/ 
|   |-- spa.traineddata                \textit{<- Base model ->}
|   |__ spa\_custom.traineddata \textit{<- Fine-tuned ->}
\end{alltt}
\end{adjustwidth}
&
\begin{adjustwidth}{0em}{0pt}
\begin{alltt}
\normalfont
PastReader/ 
|-- train/ 
|   |-- pdf/			
|   |-- ocr/         	 
|   |-- tiff/       \textit{<- tiff images (.tiff) and box files}		 
|   |__ lstmf/  \textit{<- Intermediate LSTM training files}
|-- finetuning/ 
|   |-- spa.traineddata        		
|   |__ finetuned\_model.traineddata		 
\end{alltt}
\end{adjustwidth}
\end{tabular}
\end{tcolorbox}
\caption{Comparison of the working directory structure before and after applying fine-tuning.}
\label{fig:working-dir-combined}
\end{figure}

%The automated pipeline performed the following steps: 

The fine-tuning process began with transcription alignment, where all corrected text transcriptions were renamed to adhere to the .gt.txt naming convention required by Tesseract for supervised learning. This step ensures that each transcription is correctly paired with its corresponding image during training. Next, image conversion was carried out. The original PDF files were transformed into TIFF format since TIFF images are preferable for Tesseract OCR as they provide a rasterised representation at approximately 300 DPI, which is optimal for recognition and compatible with Tesseract’s training requirements. Following this, training file generation was performed. Bounding box annotation files (.box) were created using Tesseract in training mode, capturing the spatial alignment between characters and their positions in the image. Subsequently, LSTM-compatible training data files (.lstmf) were generated by running Tesseract with the --psm 6 (assumes a block of text) and --oem 1 (uses the LSTM OCR engine) settings. These .lstmf files encapsulate the image-text pairings needed for training the recurrent neural network.

For dataset compilation, an index file named list.txt was automatically generated. This file contained the absolute paths to all .lstmf files, and served as an input manifest required by the lstmtraining binary to locate and load the training data. The model training phase was then initiated using the lstmtraining tool. The process began from an extracted LSTM file (spa.lstm), derived from the base model spa.traineddata. Training was conducted iteratively, with model checkpoints saved at regular intervals to monitor progress and facilitate recovery if needed. Finally, in the model finalisation step, the trained model weights were packaged into a usable .traineddata format using the combine\_tessdata utility. The resulting file (finetuned\_model.traineddata) represents the final fine-tuned model, ready for inference on historical document images. With this process, the updated structure of the project after fine-tuning should be as described above, in Figure \ref{fig:working-dir-combined} to the right.

To study the effect of dataset size on model performance, multiple fine-tuning experiments were conducted using incrementally larger subsets of the training data: 100, 1,000, 2,000 samples, and the complete dataset. For each configuration, a separate model was fine-tuned following the same procedure described earlier. The resulting models were then evaluated on a held-out development set to measure changes in recognition accuracy and to analyse the trade-off between increased training time and performance improvements. 

Regarding the inference process, it was applied uniformly across all models—both the baseline and each fine-tuned variant—to ensure a fair comparison. Since Tesseract requires image input, all evaluation data originally in PDF format was converted to high-resolution TIFF images, which are known to yield better OCR accuracy. Again, TIFF files were used, and they served as the input for the inference stage. Tesseract was executed via command-line interface using the same syntax for all models. The following generic command was applied: \textit{tesseract <input\_image.tiff> <output\_file> -l <model\_name> --psm 6}. This command runs Tesseract using the specified language model (the model's name without the extension .traineddata), with page segmentation mode 6 (--psm 6), which assumes a uniform block of text. This setting was selected based on empirical recommendations and its suitability for the layout of historical documents used in the dataset.

\begin{figure}[htbp]
    \centering
    \includegraphics[width=\textwidth]{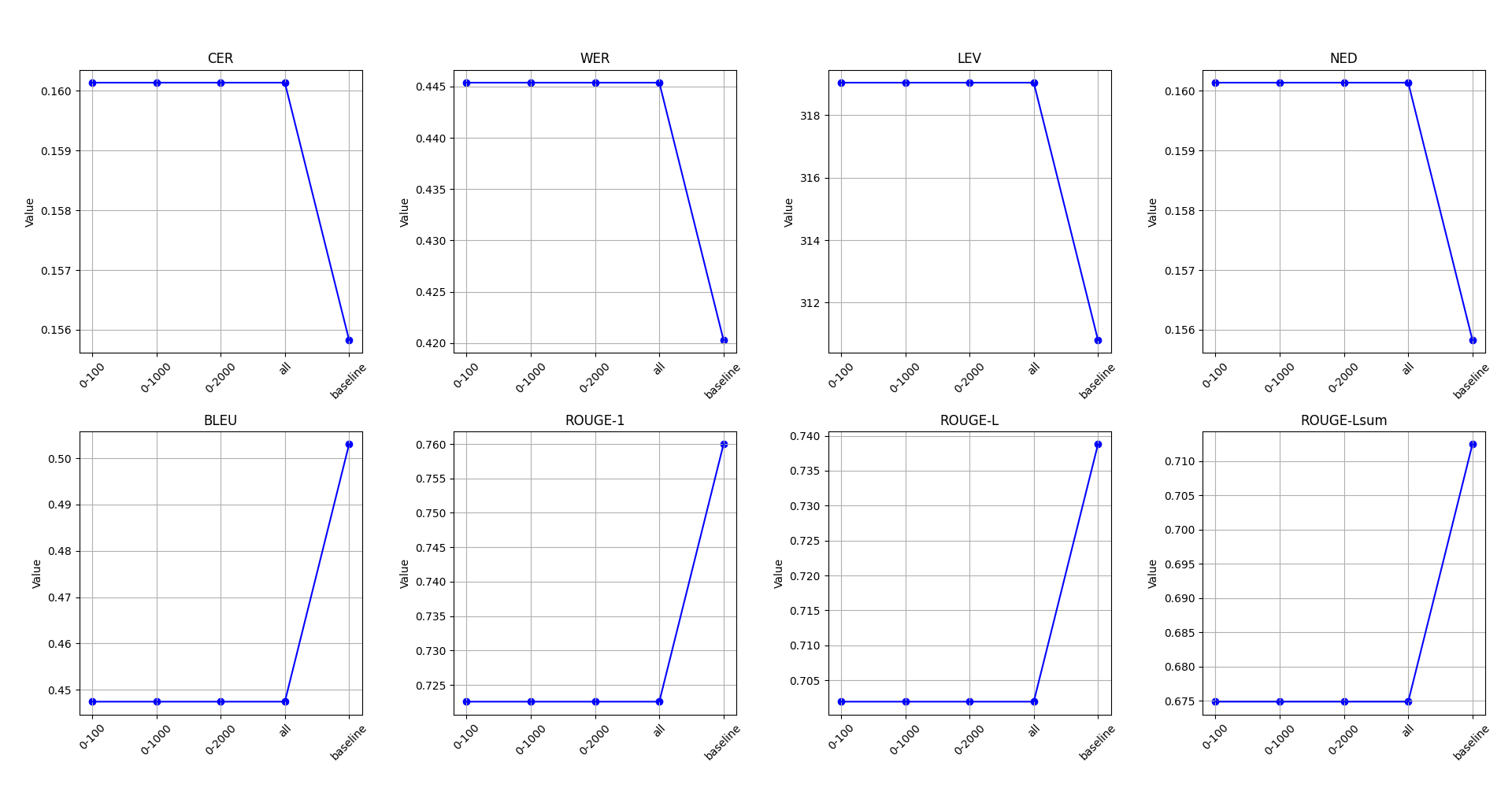}
    \caption{Tesseract charts comparing performance between fine-tuned and non-fine-tuned model using the development set.}
    \label{fig:tess}
\end{figure}

Turning now to a comparison between the performance of both Tesseract runs, Figure \ref{fig:tess} reveals a surprising trend: the baseline model (i.e., the default pre-trained model without any fine-tuning) consistently achieved the best performance across key evaluation metrics. It should be noted that these results do not correspond to the test dataset, as the training partition was used for fine-tuning and the development set for this evaluation. All fine-tuned models—regardless of the number of training samples used—produced nearly identical results, none of which surpassed the baseline established with the non-fine-tuned model. This outcome suggests that the fine-tuning process, as implemented, did not lead to meaningful performance gains and may point to limitations in the training data quality, the amount of domain-specific signal, or the need for more aggressive preprocessing or augmentation strategies. Further investigation would be required to identify the cause of this plateau and improve the efficacy of the fine-tuning approach. 

Several factors may explain the outcome. The training data was likely too heterogeneous—spanning diverse fonts, layouts, languages, and degraded document qualities—hindering the model's ability to generalise. Historical documents also deviate from the patterns Tesseract’s LSTM model is optimised for, reducing effectiveness. Misalignment between visual data and transcriptions may have introduced further noise, weakening character-level learning. Additionally, Tesseract’s sensitivity to resolution and layout demands careful preprocessing. Overall, the lack of improvement suggests that fine-tuning is only effective when training data is consistent and well-aligned.

\subsubsection{Tesseract preliminary experiment on the development set}

\begin{figure}[ht]
\centering
% Fila 1
\begin{subfigure}[t]{0.32\textwidth}
    \centering
    \includegraphics[width=\linewidth]{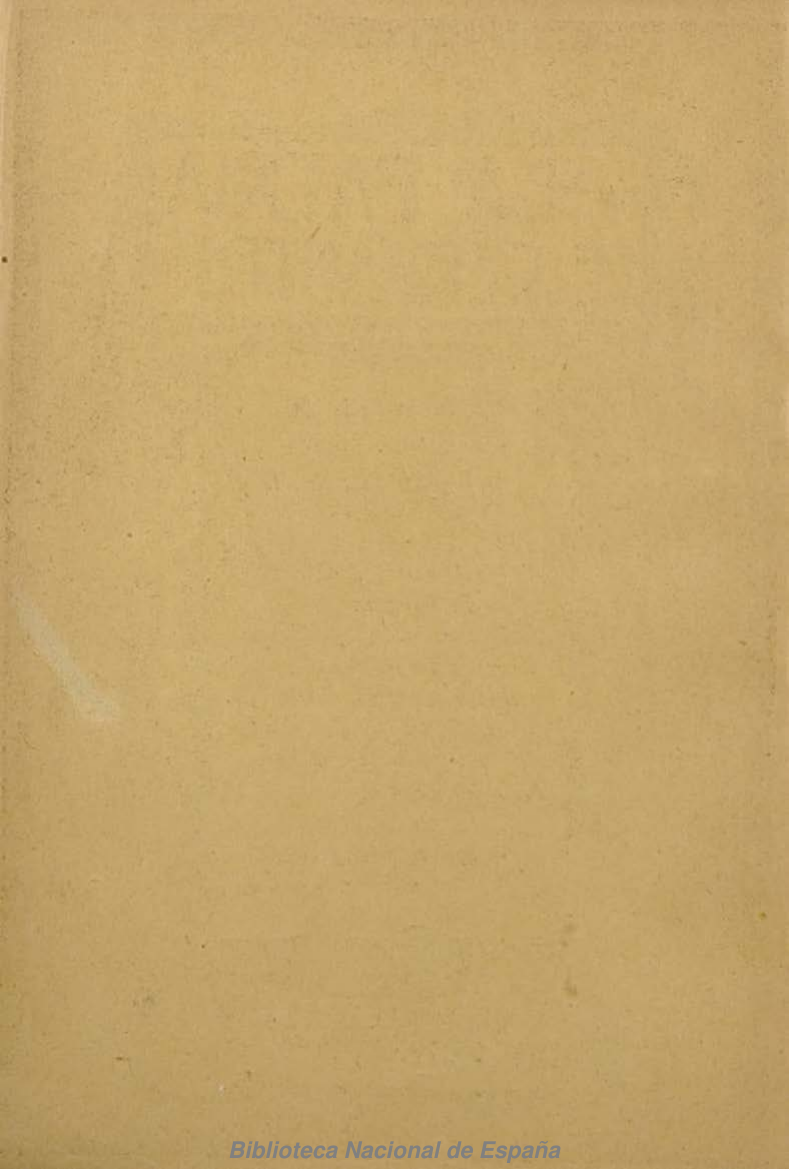}
    \caption{File 9100}
    \label{fig:9100}
\end{subfigure}
\hfill
\begin{subfigure}[t]{0.32\textwidth}
    \centering
    \includegraphics[width=\linewidth]{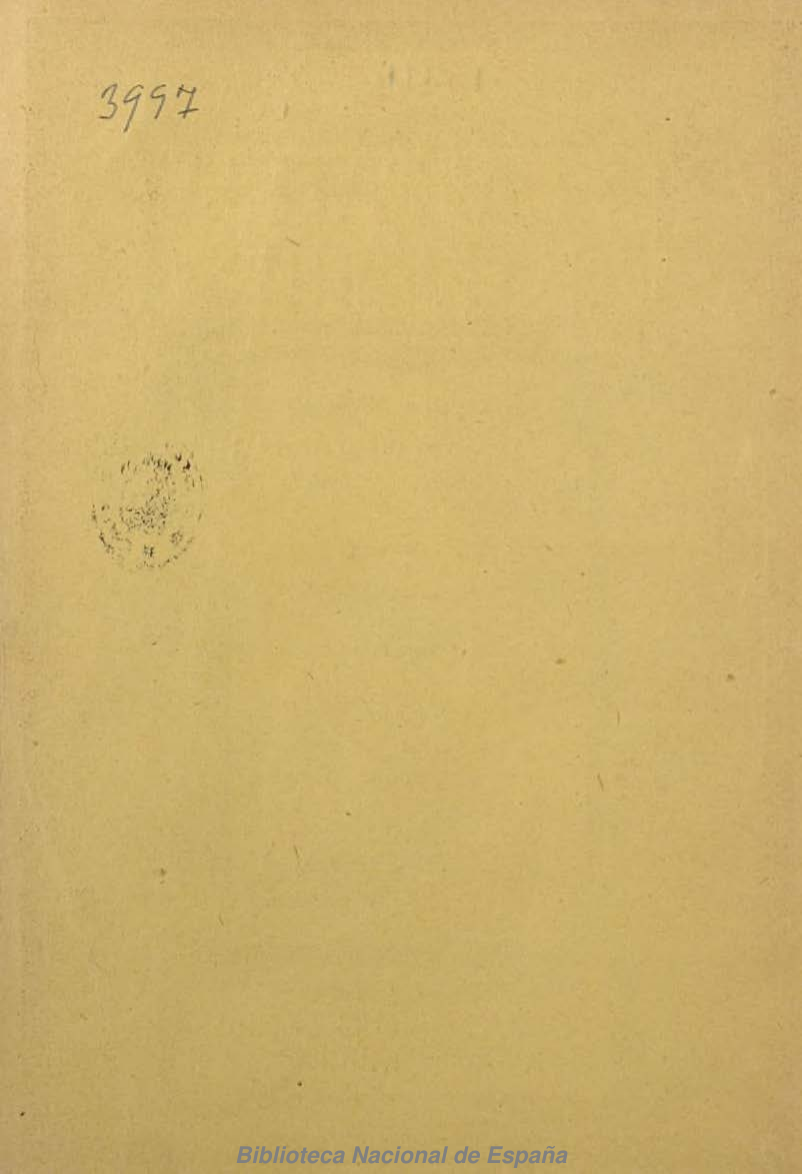}
    \caption{File 9113}
    \label{fig:9113}
\end{subfigure}
\hfill
\begin{subfigure}[t]{0.32\textwidth}
    \centering
    \includegraphics[width=\linewidth]{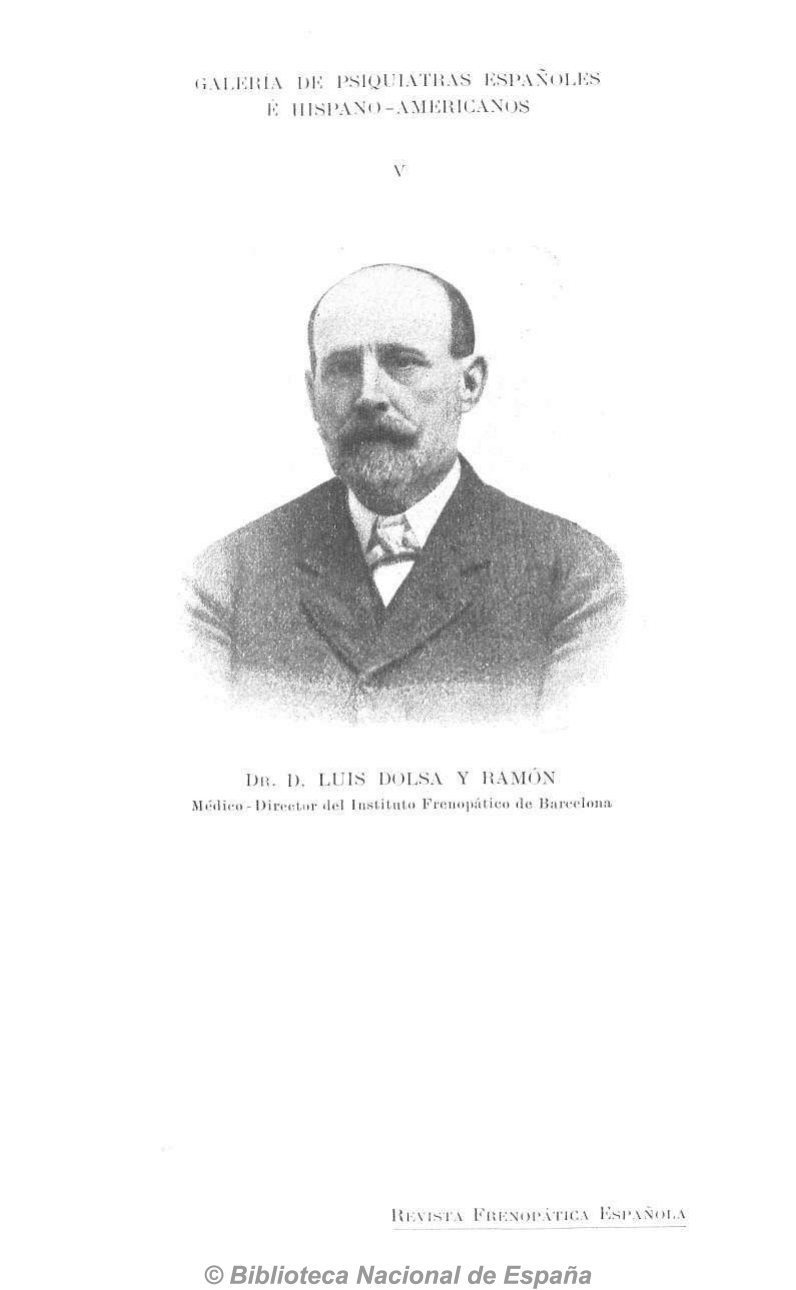}
    \caption{File 9171}
    \label{fig:9171}
\end{subfigure}

\vspace{0.5em}

% Fila 2
\begin{subfigure}[t]{0.32\textwidth}
    \centering
    \includegraphics[width=\linewidth]{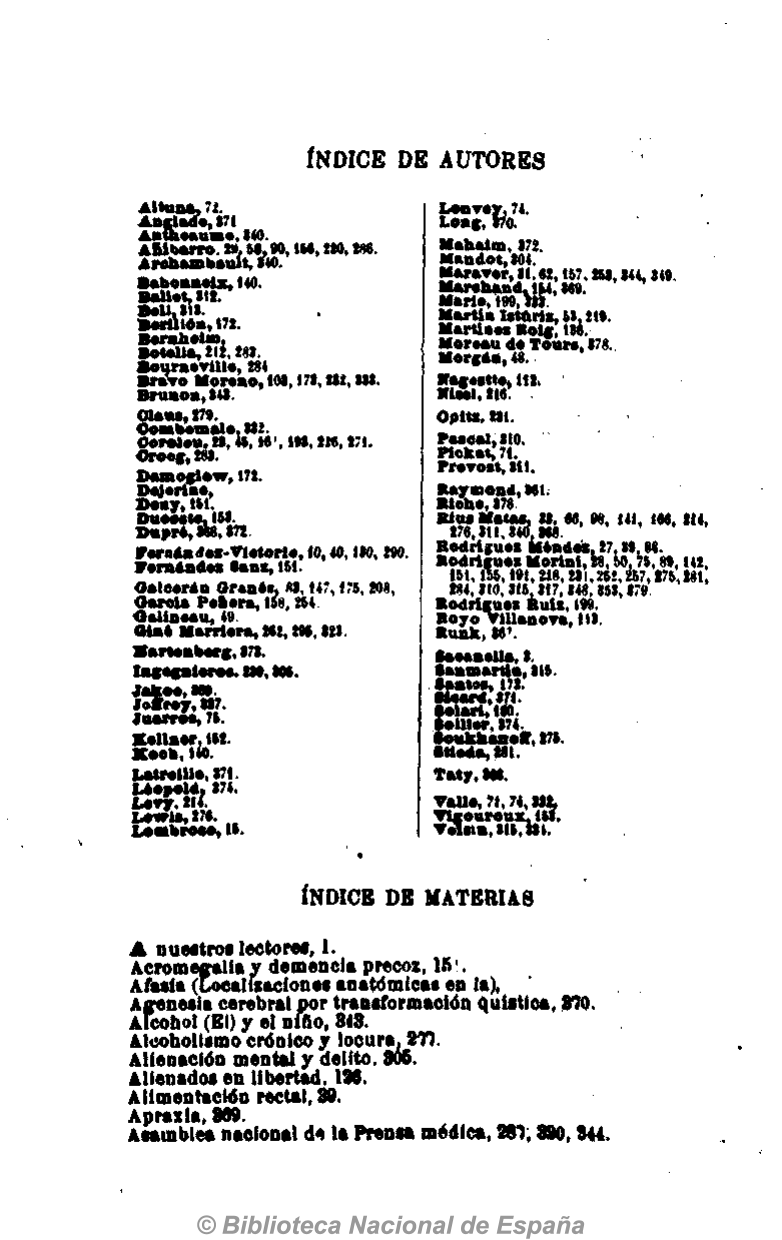}
    \caption{File 9382}
    \label{fig:9382}
\end{subfigure}
\hfill
\begin{subfigure}[t]{0.32\textwidth}
    \centering
    \includegraphics[width=\linewidth]{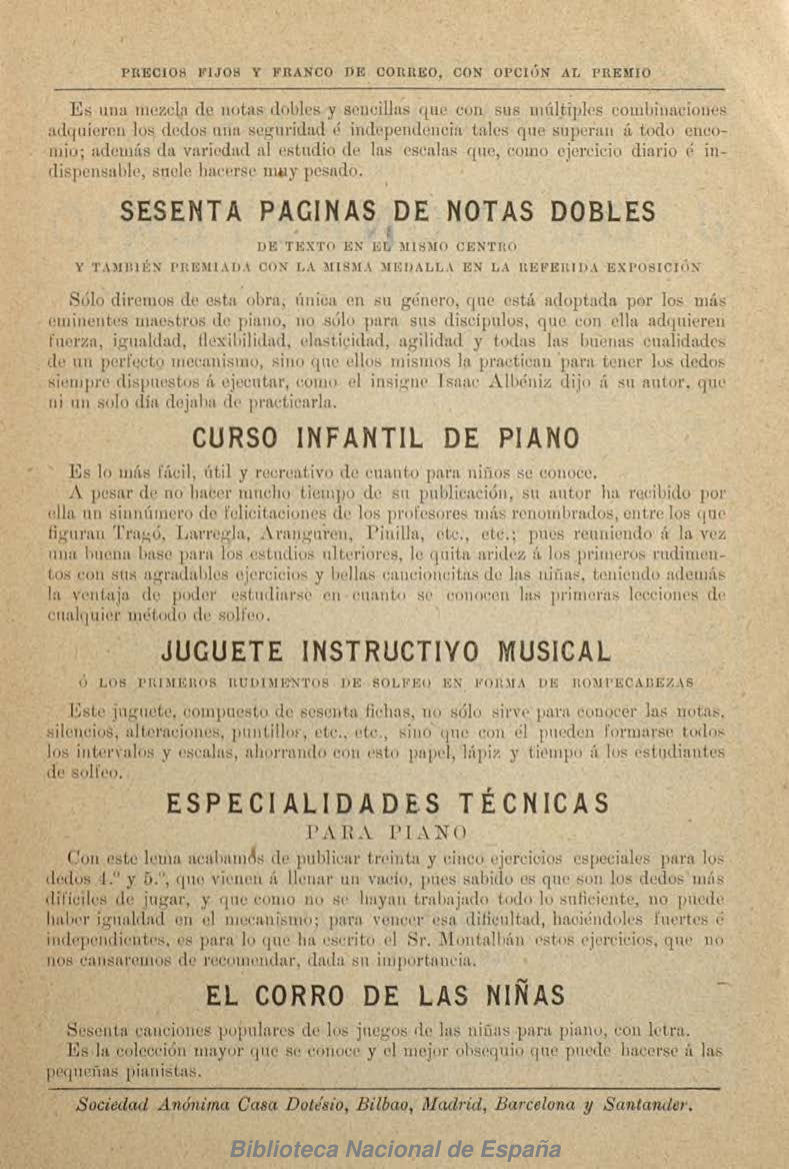}
    \caption{File 9330}
    \label{fig:9330}
\end{subfigure}
\hfill
\begin{subfigure}[t]{0.32\textwidth}
    \centering
    \includegraphics[width=\linewidth]{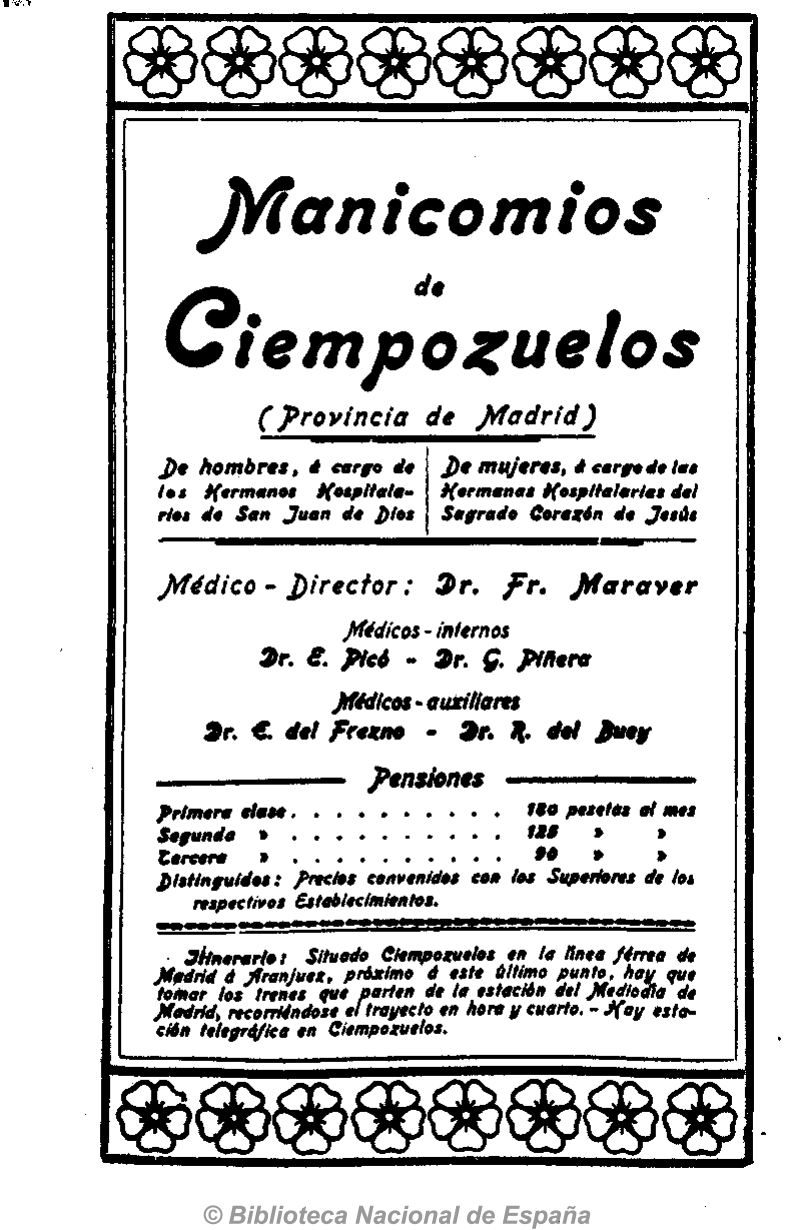}
    \caption{File 9339}
    \label{fig:9339}
\end{subfigure}

\caption{Images corresponding to the worst-performing files common to both the baseline and the fine-tuned model.}
\label{fig:pdf_examples}
\end{figure}

To gain deeper insight into the behavior of the Tesseract models, an initial qualitative analysis was conducted based on the results of running inference on the development set after fine-tuning on the training set. As a first step, it was necessary to identify the most representative samples—specifically, those for which the transcriptions yielded the poorest results. Given that the evaluation involved eight distinct metrics, the ten lowest-scoring files for each metric were initially selected. Upon comparison, it was observed that certain files appeared in multiple ``top 10'' worst-performing lists. Since this analysis was performed manually, attention was focused on the files that occurred in more than four of these lists. Because these files consistently showed poor performance across several metrics, they are assumed to be particularly informative for identifying common failure patterns and understanding the limitations of the models. In Figure \ref{fig:pdf_examples} we can observe some of the most problematic cases for both the baseline model and the fine-tuned model. 

Six files—9100, 9113, 9171, 9382, 9330, and 9399—were identified as among the worst-performing cases for both the baseline and fine-tuned models. Analysis of these files reveals several consistent failure patterns. The models tend to perform poorly on images that:
\begin{itemize}
    \item contain little or no text (e.g., 9100, 9113, 9171),
    \item exhibit low contrast (e.g., 9330), 
    \item feature non-traditional layouts resembling diagrams or schematics (e.g., 9382).  
    \item include decorative elements, as seen in file 9339.
\end{itemize}
    
Notably, while the baseline model already struggles with such inputs, the fine-tuned model additionally fails to accurately transcribe pages with decorative elements. Probably mixing somehow the information from both images and text will benefit transcription (e.g., 9100, 9113, 9171) \cite{SEPLN2025, clef-2008}. Although this analysis could be extended and refined through the examination of a larger sample, the present findings offer a preliminary understanding of the models' limitations.

\subsection{Granite3.2-vision:2b}

The remarkable ability of multimodal models to ``understand'' images and answer questions about said images makes them a perfect choice for OCR. However, without fine-tuning, i.e., only through prompting, the results are less reliable. These models are prone to analyse pictures or answering relatively complex questions about specific aspects of the image. Therefore, fine-tuning for OCR is highly advisable, mainly to generate the expected transcription exclusively, with no further explanations. Providing a benchmark for \textit{Granite3.2-vision:2b} using only prompting was our intention, but due to time constraints it was eventually impossible. However, despite inconsistencies with format and several issues, initial testing hinted promising results, especially with bigger models \textit{llama3.2-vision:11b}. Without fine-tuning, \textit{Granite3.2-vision:2b} was more prone to wrongly interpret or misspell Spanish words, probably because its training data is in English language.

We would like to give proper acknowledgments to the author of this ipynb \cite{ejemplo-granite}, as it provided a preliminary template to work with. In this example, Eli Schwartz fine-tunes \textit{Granite3.1-vision:2b}, the previous version of the model we used, using the Geometric Perception dataset \cite{geo-per-dataset} from Hugging Face, hence requiring us to adapt it to the dataset of the shared task, among other changes.

Similarly to the Tesseract approach, in order to use the model, the first step was to turn the PDF files into image files, in this case PNG files. In general terms, this kind of models use image files, so any pipeline will generally include a conversion at some point. We performed this conversion in advance for convenience, as it saves some time during processing, although at the cost of an important amount of storage.

After conversion, the dataset consists of the image of the document in PNG format, and the expected output in TXT format. However, this model requires a specific input format, similar to an interaction with a chat model. Figure \ref{fig:dataset} depicts the structure of the dataset after the input format conversion. As it can bee seen, for each example, it mainly requires a \textbf{system prompt}, an \textbf{image}, a message from the user, which is defined as \textbf{user prompt}, and what we expect from the model: the \textbf{ocr text}. The user prompt is: ``Please perform OCR on this Spanish document.'' This is the structure of the dataset, which is passed to the model with the values of the parameters according to each example.

\begin{figure}[htbp]
\centering
\begin{tcolorbox}[
    colback=white,
    colframe=black,
    boxrule=0.5pt,
    width=\textwidth,
    fontupper=\ttfamily,
    title=DATASET STRUCTURE
]

\begin{adjustwidth}{1em}{0pt}
\begin{alltt}
    chat = [
        {"role": "system", "content": [{"type": "text", "text": SYSTEM_PROMPT}]},
        {"role": "user", "content": [
            {"type": "image", "image": image},
            {"type": "text", "text": USER_PROMPT}
        ]},
        {"role": "assistant", "content": [{"type": "text", "text": ocr_text}]},
    ]
    examples.append(chat)
\end{alltt}
\end{adjustwidth}

\end{tcolorbox}
\caption{Dataset structure required by \textit{Granite3.2-vision:2b}.}
\label{fig:dataset}
\end{figure}

As mentioned, it also requires a system prompt. It is more detailed and contains precise instructions of what is expected from the model. The system prompt used can be seen in Figure \ref{fig:system_prompt} below. It instructs the model not to invent information or modify the text, while trying to obtain the raw text with no further additions.

\begin{figure}[htbp]
\centering
\begin{tcolorbox}[
    colback=white,
    colframe=black,
    boxrule=0.5pt,
    width=\textwidth,
    fontupper=\ttfamily,
    title=SYSTEM\_PROMPT
]
\begin{adjustwidth}{2em}{0pt}
\begin{alltt}
\itshape
You are an OCR expert specialised in Spanish documents.
You are analysing an old book scan with potentially low quality.

INSTRUCTIONS:

    Extract ALL text exactly as it appears.

    Do not correct, interpret or modify the text in any way.

    Return ONLY the raw text, without any additional comments or formatting.

    Do not invent content not present in the image.

The output must be EXACTLY the recognised text, without adding anything else.
\end{alltt}
\end{adjustwidth}
\end{tcolorbox}
\caption{Prompt for OCR with Spanish documents.}
\label{fig:system_prompt}
\end{figure}

It is worth noting that a challenging aspect was the use of RAM memory for loading the dataset image files as pixel values. The employed gaming pc features 32gb of RAM memory. Despite the fact that this is not a low amount of RAM for consumer-grade hardware, it falls short when attempting to process all the images at full resolution from the dataset, running out of memory. Said resolution could vary depending on the file, within the ranges of 1000–1500 pixels by 2000–3000 pixels. Even at a resolution close to 827x1169 pixels, the same error persisted. Considering that \textit{Granite3.2-vision:2b}'s processor scans images by cropping them and analysing areas of 384x384 pixels, we reduced the resolution of the dataset images to a similar size: 414x585 pixels. Using multiples of the crop area dimensions might have been more efficient. It is worth mentioning that, since the images in the dataset have slightly different aspect ratios, some additional measures had to be taken to ensure all images had the same dimensions for fine-tuning. This posed the challenge of potential information loss when resizing them to a fixed resolution given that this process usually crops the images. To address this, each image was resized while preserving its original aspect ratio, ensuring it fitted within the 414×585 pixel target. The remaining space was then filled with padding to reach the exact required dimensions. There are less hardware-demanding alternatives to loading images as pixel values, such as including image paths in the dataset instead of the actual image data and loading them during fine-tuning, or converting the pixel values to tensors ahead of time for later use. However, these options were not explored in this paper.

With respect to quantisation to reduce memory consumption, it is a cornerstone for fine-tuning on consumer-grade hardware. The model was fine-tuned using QLoRA (Quantized Low-Rank Adapter) \cite{qlora}, an approach that combines 4-bit quantisation with parameter-efficient fine-tuning based on LoRA (Low-Rank Adaptation of Large Language Models) \cite{lora} adapters. The base model was loaded in 4-bit precision using the NF4 (Normalised Float 4) quantisation scheme, along with double quantisation and bfloat16 computation to optimise numerical stability and performance. Specific modules, such as the vision tower and output head, were excluded from quantisation to preserve their full precision. Lightweight LoRA adapters were injected into the projection layers of the language model, and only these adapters were updated during training. This configuration was fundamental for fitting the model within the RTX 5080's 16gb of VRAM, although it trades off some performance.

Regarding hyperparameters, this aspect remained largely unexplored, as time limitations prevented any meaningful experimentation. The configuration was mostly based on default or initial guesses, with minimal testing. Despite the use of quantisation strategies, special care was still required. Since the task involves images, VRAM usage was inherently higher. The hyperparameters used are shown in Table \ref{tab:hyper}. As observed, the per-device batch size is set to a minimal value of 1, which is partially mitigated by using 8 gradient accumulation steps. This effectively simulates a larger batch size without exceeding memory limits. In addition, \texttt{bfloat16} precision is used alongside gradient checkpointing to further reduce memory usage during training. The fused \texttt{adamw\_torch\_fused} optimiser was selected to maximise computational efficiency. A conservative learning rate of \texttt{1e-4} and a weight decay of 0.01 were also applied to ensure stable fine-tuning. Finally, checkpoint saving was limited to the most recent state to save disk space and simplify checkpoint management.

\begin{table}[htbp]
\centering
\begin{tabularx}{\textwidth}{>{\raggedright\arraybackslash}X>{\raggedright\arraybackslash}X}
\toprule
\textbf{Hyperparameter} & \textbf{Value} \\
\midrule
Number of training epochs & 1 \\
Per-device batch size & 1 \\
Gradient accumulation steps & 8 \\
Warmup steps & 10 \\
Learning rate & \texttt{1e-4} \\
Weight decay & 0.01 \\
Logging steps & 10 \\
Save strategy & \texttt{steps} \\
Save steps & 20 \\
Save total limit & 1 \\
Optimizer & adamw\_torch\_fused \\
bfloat16 precision & True \\
Remove unused dataset columns & False \\
Gradient checkpointing & True \\
Dataset text field & "" \\
Skip dataset preparation & True \\
\bottomrule
\end{tabularx}
\caption{Hyperparameters used during fine-tuning.}
\label{tab:hyper}
\end{table}

Even with all the measures and strategies taken to fine-tune \textit{Granite3.2-vision:2b}, it narrowly fitted in the VRAM. The training dataset, comprising both the development and training sets, was randomly split into two parts: 90\% for training and 10\% for testing. After successfully fine-tuning on the training dataset, the model was used for inference on the final test dataset, made available later by the organisation. Apart from requiring the same dataset format adaptation into a chat format, there was no other significant step worth mentioning. This Granite approach resulted in \textbf{GRESEL2\_run1}.

\section{Analysis of results}
%\section{Analysis of results: test set }

\subsection{Quantitative analysis}

\begin{comment}
\textcolor{magenta}{NOTA de Yanko: No hacemos sección de cuantitativo y cualitativo. Todo junto.
Se comparan las métricas entre los distintos sistemas que hemos hecho, e intentamos ver por qué unos sacan mejor en una que en otra.
Se detectan los patrones y tendencias de lo que hace cada sistema y se analiza. Esto ayuda mucho a explicar por qué gana uno u otro y permite ver puntos de mejora.
Dedicar un párrafo o sección a la huella de carbono y energía, ya que lo tenemos, se menciona un poco.
*Transkribus también NOTA: Ya está en la sección anerior, fueron casi 3h de transcripción, pero no sabemos en qué equipo...
Poner lo de los créditos y el tiempo de ejecución. Pequeño comentario.}
\end{comment}

In order to understand Table \ref{tab:ocr-results}, a short description of each of our runs is provided below in Table \ref{tab:run-descriptions}.

\begin{table}[ht]
\centering
\begin{tabular}{|l|l|}
\toprule
\textbf{Run ID} & \textbf{Description} \\
\midrule
GRESEL1\_run1 & Using Transkribus to extract text from PDF files, preserving line breaks. \\
GRESEL1\_run2 & Using Tesseract without fine-tuning; serves as a baseline. \\
GRESEL1\_run3 & Using the fine-tuned version of Tesseract for inference. \\
GRESEL2\_run1 & Using the fine-tuned \textit{Granite3.2-vision:2b} model for inference. \\
GRESEL2\_run2 & Using Transkribus to extract text from PDFs, with joined lines (no line breaks). \\
\bottomrule
\end{tabular}
\caption{Description of the submitted runs}
\label{tab:run-descriptions}
\end{table}

Table \ref{tab:ocr-results} offers a chart with the results of the shared task provided by the organisation. Originally, six different metrics were planned to be implemented in the shared task: Word Error Rate (WER), Sentence Error Rate (SER), Levenshtein Distance, Normalised Edit Distance (NED), BLEU (Bilingual Evaluation Understudy) and ROUGE (Recall-Oriented Understudy for Gisting Evaluation). However, SER was apparently disregarded, whereas ROUGE was extended to include four variants—ROUGE1, ROUGE2, ROUGEL and ROUGELSUM—, and both WER and Levenshtein Distance were preserved, totalling eight metrics. Consequently, the OCR evaluation relies on a diverse set of complementary metrics that capture different dimensions of performance. 

\begin{figure}[htbp]
    \centering
    \includegraphics[width=\textwidth]{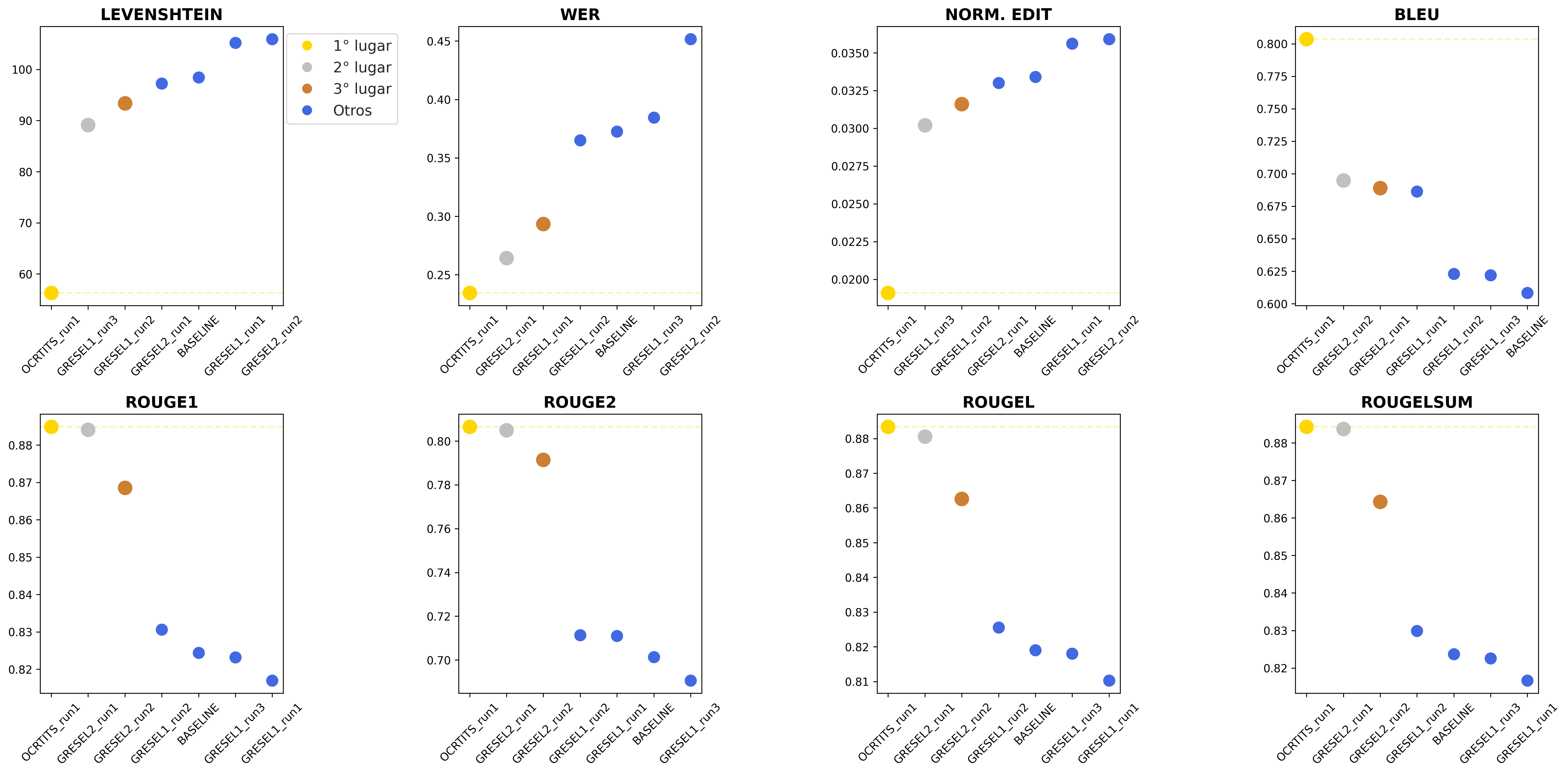}
    \caption{Scattered plots for each metric sorted by best scores.}
    \label{fig:metrics-sorted}
\end{figure}

Moreover, literal accuracy is assessed through character-level metrics such as Levenshtein Distance and NED, which count the number of required edit operations; and WER, which quantifies lexical mismatches. Therefore, these metrics favour exact textual reproduction, where lower values indicate better performance. In contrast, semantic quality is measured using BLEU, which evaluates n-gram overlap; and the ROUGE family of metrics: ROUGE-1 focuses on term-level recall, ROUGE-2 captures short-range contextual relationships, and ROUGE-L/ROUGE-LSum assess overall discourse coherence. In these metrics, higher values indicate better preservation of linguistic meaning, even when character-level differences are present. Together, these metrics provide a holistic view of system performance, from raw text fidelity to higher-level comprehension. Figure \ref{fig:metrics-sorted} provides a more visual insight into the results.

\begin{table}[ht]
\centering
\resizebox{\textwidth}{!}{%
\begin{tabular}{lrrrrrrrr}
\toprule
\textbf{TEAM} & \textbf{LEVENSHTEIN} & \textbf{WER} & \textbf{NED} & \textbf{BLEU} & \textbf{ROUGE1} & \textbf{ROUGE2} & \textbf{ROUGEL} & \textbf{ROUGELSUM} \\
\midrule
OCRTITS\_run1   & 56.3023 & 0.2344 & 0.0191 & 0.8035 & 0.8849 & 0.8065 & 0.8834 & 0.8843 \\
GRESEL1\_run3   & 89.1427 & 0.3846 & 0.0302 & 0.6220 & 0.8232 & 0.6907 & 0.8180 & 0.8226 \\
GRESEL1\_run2   & 93.3819 & 0.3650 & 0.0316 & 0.6229 & 0.8306 & 0.7114 & 0.8256 & 0.8299 \\
GRESEL2\_run1   & 97.2399 & 0.2643 & 0.0330 & 0.6890 & 0.8841 & 0.8049 & 0.8806 & 0.8837 \\
BASELINE        & 98.4497 & 0.3725 & 0.0334 & 0.6083 & 0.8244 & 0.7014 & 0.8190 & 0.8237 \\
GRESEL1\_run1   & 105.1823 & 0.2933 & 0.0356 & 0.6863 & 0.8170 & 0.7110 & 0.8103 & 0.8167 \\
GRESEL2\_run2   & 105.9065 & 0.4516 & 0.0359 & 0.6949 & 0.8686 & 0.7914 & 0.8626 & 0.8643 \\
\bottomrule
\end{tabular}%
}
\caption{Official leaderboard}
\label{tab:ocr-results}
\end{table}

Table \ref{tab:ocr-results} reveals a decent performance in all our runs overall, comprising GRESEL1 and GRESEL2 teams' runs. Most of them are above the baseline across the different metrics, which is a good indicator of minimal performance. As our approaches have been very varied it is worth analysing each approach. Following the same order, Transkribus approaches, runs \textbf{GRESEL1\_run1} and \textbf{GRESEL2\_run2}, show a predictable trend: Transkribus preserving break lines (GRESEL1\_run1) performs better in precision metrics such as Levenshtein, WER and NED compared to Transkribus joining up break lines (GRESEL2\_run2). This is a normal consequence of preserving a more similar format to the ground truth, which preserves break lines from the PDF files in the outputs. Conversely, Transkribus joining up lines performed better in the semantic metrics, BLUE and ROUGE variants, even achieving second place in BLUE. These two runs were identical except for the break-line strategies, which is extremely interesting as the notable differences in metrics perfectly showcase the criteria of each of them.

The Tesseract-based configurations—\textbf{GRESEL1\_run2} and \textbf{GRESEL1\_run3}—exhibit performance patterns that align with prior results obtained during training-phase experiments (refer to Figure \ref{fig:tess}). Notably, the non-fine-tuned model (GRESEL1\_run2) outperformed the fine-tuned counterpart in six out of eight evaluated metrics, showing broader robustness across both surface-level and semantic evaluations. In contrast, the fine-tuned model (GRESEL1\_run3) attained higher scores in only two precision-focused metrics: Levenshtein distance, where it secured the second-best overall score, and normalised edit distance (NED). These results point toward a potential overfitting effect during the fine-tuning process, wherein the model may have adapted too closely to specific features of the training data and failed to generalise effectively to unseen test samples. Nonetheless, performance gaps in semantic metrics such as ROUGE-L (0.8180 for GRESEL1\_run3 vs. 0.8256 for GRESEL1\_run2) and BLEU (0.6220 vs. 0.6229) remain relatively minor, indicating that fine-tuning did not drastically compromise the model's semantic coherence. Rather, it maintained competitive performance, albeit without delivering consistent improvements.

The focus now shifts to \textit{Granite3.2-vision:2b}, represented solely by its fine-tuned run, \textbf{GRESEL2\_run1}. This was arguably our strongest submission overall, depending on the relative weight assigned to each metric. The run achieved a solid second place in five out of the eight evaluation metrics, including WER and all ROUGE variants. In fact, the differences between Granite and the top-performing system \cite{narbona} in these metrics are marginal: ROUGE1 (0.8841 vs. 0.8849), ROUGE2 (0.8049 vs. 0.8065), ROUGE-L (0.8806 vs. 0.8834), ROUGE-LSUM (0.8837 vs. 0.8843), and WER (0.2643 vs. 0.2344). It also ranked third in BLEU, only behind the Transkribus run that joined up line breaks (GRESEL2\_run2) and the winner. However, Granite underperformed in character-level precision, ranking fourth in both Levenshtein Distance and NED. These results suggest that the model demonstrates excellent semantic understanding and word-level alignment with the ground truth, but has room for improvement in literal character accuracy. The observed differences between our two Transkribus runs hint that Granite may also be less effective at preserving original line breaks, although this remains a preliminary hypothesis. Given the compact size of the model and the consumer-grade hardware constraints noted earlier, the performance achieved by \textit{Granite3.2-vision:2b} is highly notable.

\begin{figure}[htbp]
    \centering
    \includegraphics[width=\textwidth]{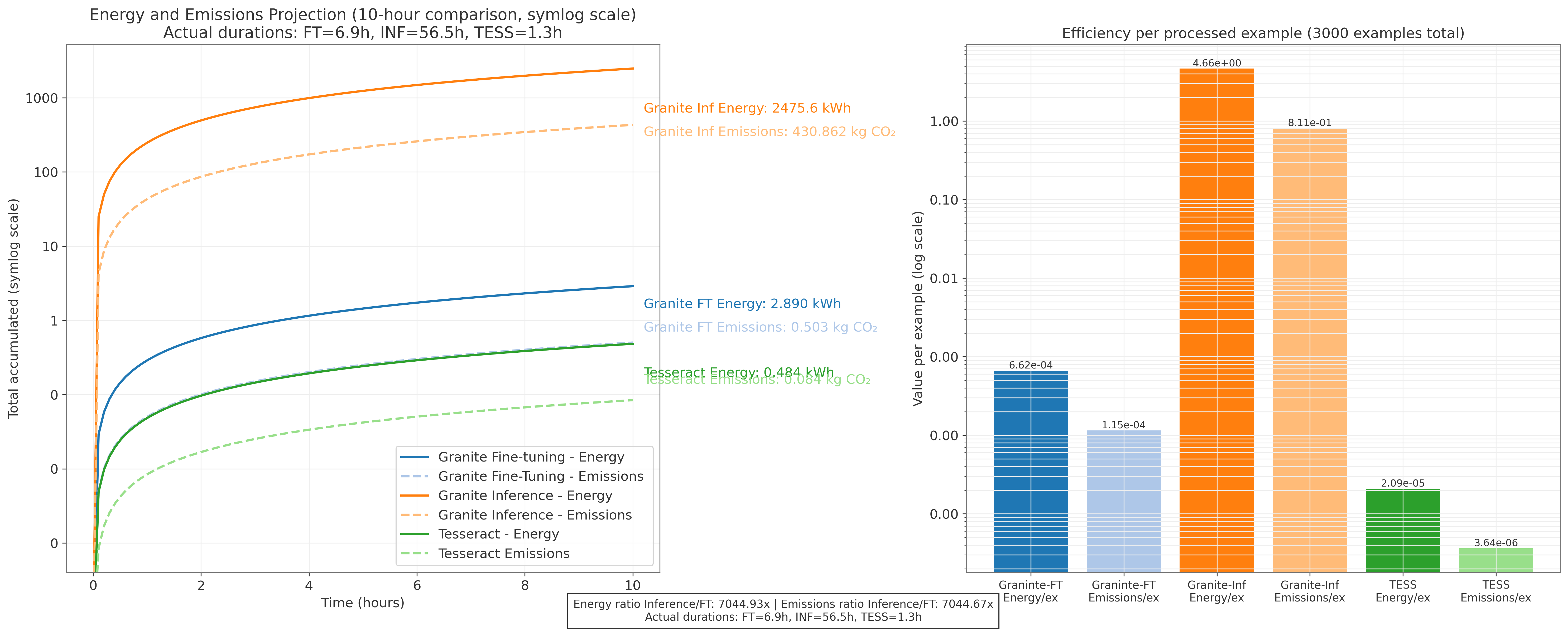}
    \caption{Energy and emission estimations: \textit{Granite3.2-vision:2b} and Tesseract.}
    \label{fig:granite-emissions}
\end{figure}

Regarding emissions, Figure \ref{fig:granite-emissions} presents a comparative analysis of computational costs. The left chart uses a logarithmic scale to visualise both Tesseract and Granite's fine-tuning and inference metrics during a simulated 10-hour execution window, enabling direct comparison of energy consumption (kWh) and CO\textsubscript{2} emissions (kg) despite their orders-of-magnitude differences. Tesseract representation includes both fine-tuning and inference. The right chart similarly uses logarithmic scaling to display per-example efficiency metrics across the 3,000-sample dataset. These charts highlight a dramatic difference between the fine-tuning process and the inference process in Granite's run. Unsurprisingly, Tesseract is significantly more efficient. In fact, a modest increase in emissions can be associated with the fine-tuned Tesseract version. While Transkribus emissions could not be precisely quantified due to its cloud-based web application architecture, the platform estimated approximately 3 hours of processing time for inference. Although two separate Transkribus runs were initiated, the inference process effectively occurred only once, as the system stores and serves results directly from its cloud infrastructure, eliminating redundant computations.

\subsection{Qualitative insights}

Upon inspecting the output generated by Granite model, it was quite satisfactory overall, but some patterns were identified. For instance, when a word is split across lines by a hyphen at the end of a line, the model joins the parts, removes the hyphen, and places the line break after the end of the reconstructed word. The model also misinterprets certain letters and words, particularly those with diacritical marks. This might be due to the fact that \textit{Granite3.2-vision:2b} was not specifically pre-trained on Spanish texts. In general, when the model is unable to recognise a difficult area, it tends to hallucinate content. It also tends to ``correct''  misspelled words found in the original, even when those words reflect historical variants of Spanish used at the time. These tendencies contribute to lower character-level accuracy, which aligns with the poor results observed in the Levenshtein Distance and NED metrics. On a positive note, the model consistently refrains from generating output when the original file is blank.

\begin{figure}[ht]
\centering
\begin{tabular}{>{\centering\arraybackslash}m{0.32\textwidth}
                >{\centering\arraybackslash}m{0.32\textwidth}
                >{\centering\arraybackslash}m{0.32\textwidth}}
    \begin{minipage}[t]{\linewidth}
        \centering
        \includegraphics[width=0.95\linewidth]{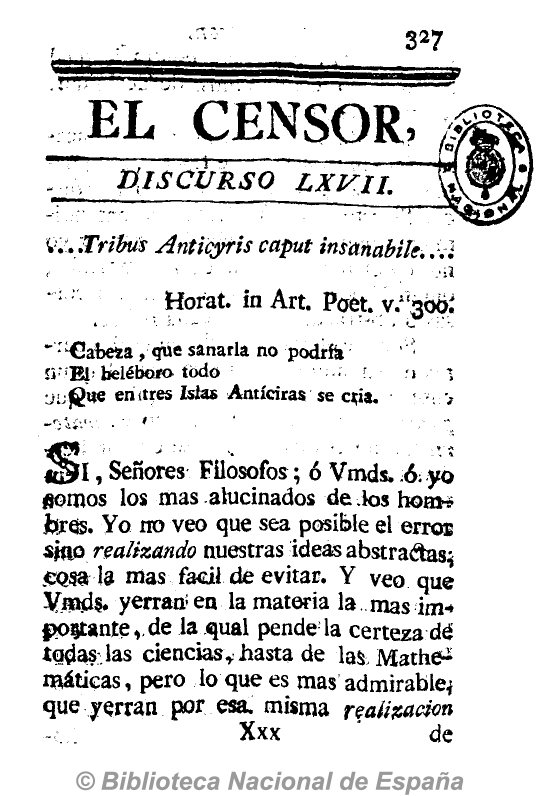}
        \captionof{figure}{File 2466}
        \label{fig:2466}
    \end{minipage} &
    \begin{minipage}[t]{\linewidth}
        \centering
        \includegraphics[width=0.95\linewidth]{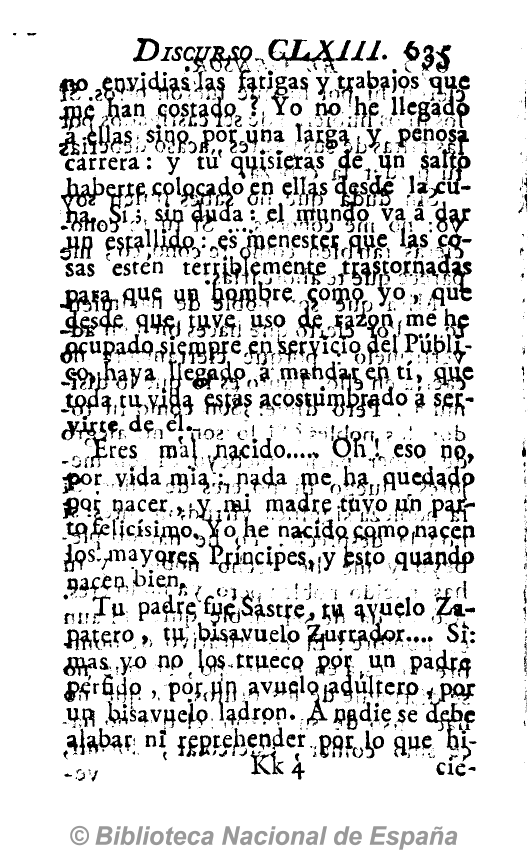}
        \captionof{figure}{File 410}
        \label{fig:410}
    \end{minipage} &
    \begin{minipage}[t]{\linewidth}
        \centering
        \includegraphics[width=0.95\linewidth]{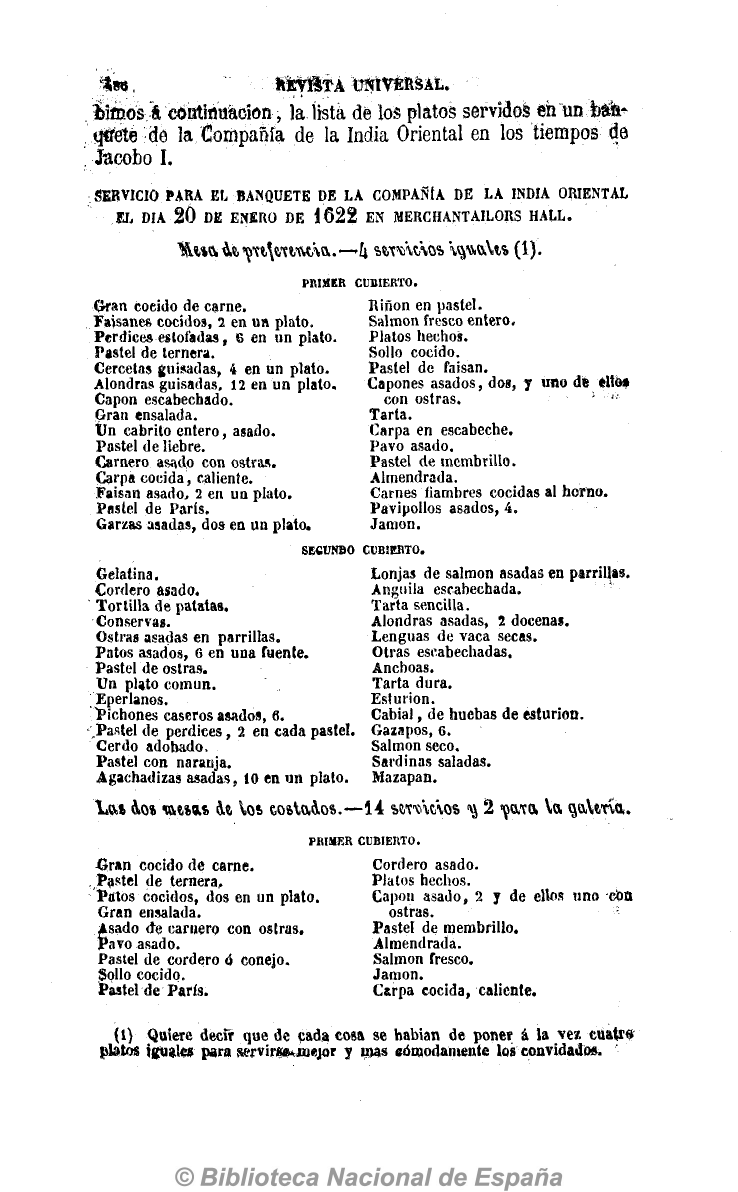}
        \captionof{figure}{File 1506}
        \label{fig:1506}
    \end{minipage} \\
\end{tabular}
\end{figure}

%``large instruction tuning dataset''
It is worth showcasing some challenging examples from the test set. The first example, shown in Figure~\ref{fig:2466} (file 2466), presents two particularly interesting elements: the stamp featuring the words \textit{BIBLIOTECA NACIONAL} (Spanish National Library), and the drop cap in the letter ``S.'' Regarding the words in the stamp, they were ignored by the model, whereas the drop cap was correctly recognised. A word misspelled in historical Spanish, ``Mathe-máticas'' (Maths), was merged and corrected to \textit{Matemáticas}. In general, the model is more prone to confuse letters or words when the area is blurry or stained.

Regarding the second example, shown in Figure~\ref{fig:410}, the image might initially appear extremely challenging, as the original physical page seems to have been stained with ink from the facing page on its right. As a result, it contains horizontally mirrored lines superimposed between the actual lines of text. Surprisingly, despite this potential challenge, the model did not struggle with these artefacts. What hindered the recognition of certain words was the presence of blurred letters or smudged areas, yet the overall transcription is quite acceptable given the circumstances.

The last example, Figure~\ref{fig:1506} displays a two-column layout. This should not pose a major challenge, as OCR models are increasingly better at handling multi-column formats. However, the transcription of this page was particularly inaccurate. The second (right-hand) column was apparently ignored, while the first (left-hand) column was heavily hallucinated. The page contains a menu, and the model invented several dishes for no apparent reason, even adding fabricated information about portion sizes in some cases. This failure to correctly render the column structure might be due to a lack of sufficient multi-column examples in the training dataset—although this remains a hypothesis.

Finally, a last observation about the dataset. While a more thorough analysis would be required to confirm this, the dev/train sets used for fine-tuning appear to be more diverse in terms of sources than the test set. The latter lacks documents containing images, while the former includes materials with different colour palettes in the paper background. Files 9062, 9171 and 8963 in Figure \ref{fig:worst_pages}, from the dev/train sets, clearly illustrate this. The test set displays less variety in page types and tends to feature whiter backgrounds, in contrast to the yellowish tones found in the dev/train documents. For example, \textit{El Censor} is one of the predominant sources in the test set (Figure \ref{fig:2466}). This mismatch between the dev/train and test sets might have negatively affected the model's learning and generalisation, although this remains a hypothesis. Ensuring both representativeness and diversity across dataset splits is crucial for effective fine-tuning.

\section{Conclusions}

This paper has explored the participation of the GRESEL team in this challenging OCR shared task. Despite the difficulties imposed by the nature and diversity of the dataset, all our runs have achieved satisfactory results compared to the baseline. We have provided a thorough analysis of the dataset from the competition, showcased how break lines affect the scores in different types of metrics, proposed several OCR approaches yielding different results and conclusions, and, ultimately, offered some ideas about how to fine-tune small multimodal models for OCR tasks and what the expected results are. It is worth highlighting the notable performance of \textit{Granite3.2-vision:2b}, mainly in semantic and word-selection metrics. Although consumer-grade hardware constraints forced the use of a smaller variant and several measures that downgraded fine-tuning quality, it excelled in performance. Naturally, some errors commonly associated with AI models were observed, such as the tendency to hallucinate words when the original text is illegible, or to ``correct'' misspelled words found in the source. Overall, this experience has enabled us to learn from our experiments and share our insights into this OCR task.

In future work, we will explore the implementation of a dataset variant from this shared task, but with transcription guidelines focusing on philological and palaeographic criteria, as anticipated. Our hypothesis is that this dataset variant will improve the model's learning. Furthermore, other similarly sized multimodal models will be employed to compare performance between models. Additionally, larger models will be explored, and full-quality fine-tuning will be applied without downgrading the quality of the data through the use of more capable hardware and cloud computing at our disposal. We expect to further enhance our results in OCR tasks by exploring new approaches and ideas.

\begin{acknowledgments}

This work is framed under the coordinated Spanish National Project GRESEL: UNED (PID2023-151280OB-C22); UAM (PID2023-151280OB-C21). It has also been partially funded by the PTA2023-023812-I grant, awarded to Yanco Amor Torterolo Orta by MICIU/AEI/10.13039/501100011033 and the European Social Fund Plus (ESF+).
  
\end{acknowledgments}

%% The declaration on generative AI comes in effect
%% https://ceur-ws.org/GenAI/Policy.html
%taxonomía: ceur-ws.org/genai-tax.html
\section*{Declaration on Generative AI}  
 During the preparation of this work, the authors used ChatGPT-4o and Deepseek-V3 in order to: Grammar and spelling check, paraphrase and reword. Besides, Microsoft's Copilot was also used in order to:  Formatting assistance (latex commands, image labelling and table creation). Further, the authors used the first two models for figures \ref{fig:metrics-sorted} and \ref{fig:granite-emissions} in order to: Generate charts based on data. After using these tools/services, the authors reviewed and edited the content as needed and take full responsibility for the publication’s content.
 
%For help with latex commands for table creation and image labelling.

%%
%% Define the bibliography file to be used
\bibliography{nuestro-paper}

\appendix

\end{document}